# An Audio-enriched BERT-based Framework for Spoken Multiple-choice Question Answering

*Chia-Chih Kuo, Shang-Bao Luo, Kuan-Yu Chen*

National Taiwan University of Science and Technology, Taiwan

{M10815022, M10615012, kychen}@mail.ntust.edu.tw

## Abstract

In a spoken multiple-choice question answering (SMCQA) task, given a passage, a question, and multiple choices all in the form of speech, the machine needs to pick the correct choice to answer the question. While the audio could contain useful cues for SMCQA, usually only the auto-transcribed text is utilized in system development. Thanks to the large-scaled pre-trained language representation models, such as the bidirectional encoder representations from transformers (BERT), systems with only auto-transcribed text can still achieve a certain level of performance. However, previous studies have evidenced that acoustic-level statistics can offset text inaccuracies caused by the automatic speech recognition systems or representation inadequacy lurking in word embedding generators, thereby making the SMCQA system robust. Along the line of research, this study concentrates on designing a BERT-based SMCQA framework, which not only inherits the advantages of contextualized language representations learned by BERT, but integrates the complementary acoustic-level information distilled from audio with the text-level information. Consequently, an audio-enriched BERT-based SMCQA framework is proposed. A series of experiments demonstrates remarkable improvements in accuracy over selected baselines and SOTA systems on a published Chinese SMCQA dataset.

## 1. Introduction

Owing to the fact that the voice assistant applications are frequently installed in a variety of mobile phones, home devices, and so on, spoken question answering (SQA) has been an emergent challenge in recent years. Apart from that, the development of multimedia technology and the popularity of video/audio sharing websites and social networks have also led to significant growth in spoken content nowadays. This has also increased the demand for machine comprehension of spoken content, which is a special case of SQA. To mitigate these challenges, a common strategy is to employ an automatic speech recognition (ASR) system to translate speech contents into text utterances. After that, a text-based method can be readily applied to the auto-transcribed text. However, it is obvious that ASR errors will inevitably affect the performance of the common strategy. Moreover, the acoustic-level information embedded in the speech may provide additional cues that is not covered by the (auto-transcribed) text. Therefore, if we can find some ways to distill and leverage the acoustic-level information, we may enhance the performance of SQA.

In this paper, we focus on the spoken multiple-choice question answering (SMCQA) task, where passages, questions, and choices are all given in speech. The major contributions are at least twofold.

First, in order to distill suitable cues from audio for compensating text-level information, a novel mechanism is proposed for acoustic-level feature encoding. Second, inspired from recent success of the bidirectional encoder representations from transformers (BERT) [1], the paper strives to develop an efficient and effective SMCQA framework based on BERT. Subsequently, the proposed framework targets at enhancing the performance of SMCQA task by assembling the contextualized language representations inferred by BERT and the acoustic-level information extracted from audio. We turn the whole process an audio-enriched BERT-based (aeBERT) framework. Evaluated on the data of "Formosa Grand Challenge – Talk to AI", a Mandarin Chinese SMCQA contest held in 2018, the proposed aeBERT framework can outperform various SOTA systems by a large margin.

## 2. Related Work

### 2.1. The Language Representation Methods

Because of the impressive successes in many NLP-related tasks, language representations have become a popular research recently. In general, the research spectrum can be classified into two main streams according to the usages for the downstream tasks [1]: (1) feature-based models and (2) fine-tuning methods. Famous and well-practiced representatives for the feature-based models are the word embedding methods. The neural network language model [2], which mainly concentrates on estimating an *n*-gram language model while inducing word embeddings as a by-product, is the most-known seminal study on developing various word embedding methods. Such an attempt has already motivated many follow-up extensions to develop similar methods for probing latent semantic and syntactic regularities in the representation of words. The inferred word embeddings are usually treated as feature vectors for downstream tasks. Representative methods include, but are not limited to, the CBOW [3], the skipgram [3], the GloVe [4], and the ELMo [5]. On the contrary, the OpenAI GPT [6], the BERT [1], the XLNet [7], the RoBERTa [8], and the ALBERT [9] are the leading methods in the latter class. The fine-tuning methods usually consist of two parts: pretraining and task-specific parameters finetuning. Formally, such a school of methods usually leverage an unsupervised objective to obtain a pretrained model, and then they introduce minimal task-specific parameters and train on the downstream task by simply finetuning all the (or only the task-specific) parameters [6]. Recently, the latter category (i.e., finetuning approaches) becomes a dominative research subject in NLP community.

### 2.2. The Multiple-choice Question Answering

In a text-based MCQA task [10–14], the input to the model includes a passage, a question, and several answer choices. The passage usually consists of several sentences, while the question and each answer choice are a single sentence. The question answering model is designed to select a correct answer from multiple choices based on the information given in the passage and question. Previous studies usually concentrated on utilizing lexical and syntactic information in the given texts to infer the answer [15–18], while recent research has turned to present various MCQA models based on neural networks [11–14]. Classic methods include the hierarchical attention-based CNN [19], the parallel-hierarchical neural model [20], and the hierarchical attention flow model [21], to name just a few. Among them, the query-based attention CNN (QACNN) model [13] is a representative.

Opposite to the conventional MCQA task, the passage, question, and choices are all in speech in a spoken MCQA task. A naïve but easy solution is to first transcribe these speech utterances into text using an ASR system. Thereafter, a text-based method (e.g., QACNN) can be readily applied to the auto-transcribed text. Such a strategy only considers text-level information, but it is obvious that the audio may contain useful cues for answer prediction. Hence, several studies have been proposed to cope with the SMCQA task by considering both text-level and acoustic-level features. The CNN-based hierarchical multistage mutimodal (HMM) framework [22] and the SpeechBERT [23] model are representatives. The former tries to explore both the text-level and the acoustic-level relationships between a pair of passage and choice as well as a pair of passage and question by CNN-based attention mechanism. The latter assumes that the passage is given in the form of speech, while the question is in the form of manual transcription (i.e., without recognition errors). A concatenation of question and passage can then be fed into the BERT model, which makes it possible to explore the relationship between acoustic-level and text-level cues with the self-attention [24] mechanism. Although the HMM model seems to equip comprehensive ability for SMCQA task, it doesn't leverage the merits in recent language representation models (e.g., BERT). SpeechBERT, which takes acoustic features as additional inputs to the BERT model, may suffer from the input length limitation of BERT so as to downgrade the performance and make the model inflexible.

## 3. The Methodology

### 3.1. The Vanilla BERT Method

Recently, among the popular language representation methods, BERT [1] has attracted much interest due to its state-of-the-art performances in several NLP-related tasks. A naïve but efficient way is to employ BERT as an input encoder, and then a set of simple and task-specific layers is stacked upon the BERT model. Subsequently, the additional task-specific parameters can be finetuned toward to optimize the performance of the target task [25–27]. When BERT comes to the SMCQA task, a straightforward strategy is to be employed to encode a concatenation auto-transcribed token (i.e., wordpiece) sequence of a passage, a question, and one of choices, and then a simple classification objective is introduced to indicate which choice is the correct answer to the question according to the given passage. More formally, for a passage $p = \{w_1^p, w_2^p, \ldots, w_{|p|}^p\}$, a question $q = \{w_1^q, w_2^q, \ldots, w_{|q|}^q\}$ and $n^{th}$ choice $c_n = \{w_1^{c_n}, w_2^{c_n}, \ldots, w_{|c_n|}^{c_n}\}$, a concatenation token sequence $\{[CLS], w_1^q, \ldots, w_{|q|}^q, w_1^{c_n}, \ldots, w_{|c_n|}^{c_n}, [SEP], w_1^p, \ldots, w_{|p|}^p, [SEP]\}$ can be obtained, where "[CLS]" represents a special token of every concatenation word sequence and "[SEP]" is a separator token. Next, the pretrained BERT model is used to extract a set of hidden vectors for each token in the concatenation token sequence. In order to select an answer to the question, a single layer neural network is adopted to translate the hidden vector corresponding to the "[CLS]" token to a score. After that, a softmax function is stacked upon all of the scores of candidate choices [14]. Consequently, the objective of the downstream SMCQA task is to maximize the likelihood of the correct choices in the training examplers. We term the model as "vanilla BERT". Although the naïve strategy can already obtain a certain level of performance, the acoustic-level information is ignored in vanilla BERT model and the recognition errors may mislead the final predictions so as to downgrade the performance. Accordingly, an audio-enriched BERT-based framework is presented in this paper to further boost the performance of SMCQA.

### 3.2. The Audio-enriched BERT-based Framework

Previous studies have evidenced that acoustic-level information can be complementary materials to the auto-transcribed text for choosing answers in the SMCQA task, thus an essential challenge is to design a theoretical and systematic way to incorporate the acoustic cues with the text-level information, which is readily well-modeled by BERT. To achieve the goal, for each auto-transcribed token $w_i$ and its corresponding acoustic feature vectors (e.g., MFCCs) $\mathcal{F}^{w_i} = \{f_1^{w_i}, \cdots, f_{|w_i|}^{w_i}\} \in \mathbb{R}^{d_a \times |w_i|}$, we intend to learn an attention map, which can reveal the importance degree of each feature vector and acoustic statistic:

$$A^{w_i} = \text{softmax}(\mathcal{W}_a \mathcal{F}^{w_i}) \quad (1)$$

where $\mathcal{W}_a \in \mathbb{R}^{d_a \times d_a}$ is a learnable model parameter, $A^{w_i} \in \mathbb{R}^{d_a \times |w_i|}$ is the resulting attention map, $d_a$ denotes the dimension of the acoustic feature vector, and $|w_i|$ is referred to the number of acoustic feature vectors for token $w_i$. Accordingly, an acoustic-level representation $v^{w_i} \in \mathbb{R}^{d_a}$ for $w_i$ can be obtained by:

$$v^{w_i} = \sum_{j=1}^{|w_i|} [A^{w_i} \odot \mathcal{F}^{w_i}]_j \quad (2)$$

where $\odot$ means element-wise product of matrices, and $[\cdot]_j$ denotes $j$-th column of a matrix. It is worthwhile to indicate that the inferred attention map considers temporal and spectral statistics simultaneously, thus it can encapsulate a set of acoustic feature vectors into a single vector representation from a holistic perspective. Additionally, in our preliminary experiments, the proposed strategy can give superior results than conventional attention methods, such as vector-wise self-attention mechanism [24]. We term the process a temporal-spectral attention mechanism and denote hereafter as TSAtt in short.

Next, to inherit the advantages of the BERT model and leverage the inferred elaborative acoustic-level representations by TSAtt, a straightforward yet effective approach that appends all of the information cues to the embeddings of the input tokens is proposed. Figure 1 illustrates the overview of the proposed framework. Again, for a passage $p = \{w_1^p, w_2^p, \ldots, w_{|p|}^p\}$, a question $q = \{w_1^q, w_2^q, \ldots, w_{|q|}^q\}$ and $n^{th}$ choice $c_n = \{w_1^{c_n}, w_2^{c_n}, \ldots, w_{|c_n|}^{c_n}\}$, a concatenation token sequence $\{[CLS], w_1^q, \ldots, w_{|q|}^q, w_1^{c_n}, \ldots, w_{|c_n|}^{c_n}, [SEP], w_1^p, \ldots, w_{|p|}^p, [SEP]\}$ can be obtained. Each token $w_i$ in the input sequence can map to a composite vector by summing its corresponding token embedding, position embedding, and segment embedding as conventional BERT model except an acoustic-level representation $\hat{v}^{w_i}$:

$$\hat{v}^{w_i} = \mathcal{W}_s v^{w_i} + b_s \quad (3)$$

where $\hat{v}^{w_i} \in \mathbb{R}^{d_t}$ is a simple variant of $v^{w_i}$ by passing through a fully-connected feed-forward layer with parameters $\mathcal{W}_s \in \mathbb{R}^{d_t \times d_a}$ and $b_s \in \mathbb{R}^{d_t}$. The simple layer is mainly used to align the dimension of original acoustic-level representation $v^{w_i}$ with other embeddings. In other words, compared to the vanilla BERT model, an extra acoustic-level embedding $\hat{v}^{w_i}$ is included in addition to the conventional embeddings (i.e., word embedding, segment embedding and positional embedding) for each token. By doing so, the proposed strategy introduces the acoustic-level information to BERT in a natural way. After that, BERT is employed to infer contextualized representations for each token in the sequence as usual, and the resulting vector for the "[CLS]" token can be viewed as a comprehensive representation $h^{[CLS]^{c_n}}$, which can then be used to determine a relevance score $r^{c_n}$ by a fully-connected feed-forward network for choice $c_n$:

$$r^{c_n} = \mathcal{W}_r h^{[CLS]^{c_n}} + b_r \qquad (4)$$

where $\mathcal{W}_r \in \mathbb{R}^{1 \times d_t}$ and $b_r \in \mathbb{R}$ are the parameters of the fully-connected feed-forward layer. Finally, in the training stage, a softmax function is stacked upon all of the scores of candidate choices for each pair of question and passage:

$$P(c_n) = \frac{exp(r^{c_n})}{\sum_{n'=1}^{N} exp(r^{c_{n'}})} \qquad (5)$$

where $N$ denotes the number of candidate choices. Subsequently, the training objective of the downstream SMCQA task is to maximize the likelihood of the correct choices in the training examplers. For testing, the choice with the largest relevance score (i.e., $r^{c_n}$) will be selected as the answer without performing the nonlinear normalization.

In a nutshell, the proposed framework composes of two key ideas. First, a TSAtt mechanism is proposed to condense a set of acoustic feature vectors into a meticulous vector representation. Second, a simple but natural strategy, which modifies the embedding information for a token by injecting the acoustic-level cues and thus has influence on the token representation generated by BERT, not only for this token but also for other tokens since BERT considers contextual information, is proposed. We turn the whole process an audio-enriched BERT-based framework (aeBERT) hereafter.

## 4. Experimental Setup

### 4.1. Dataset

We evaluated the proposed aeBERT framework on the "2018 Formosa Grand Challenge – Talk to AI[1]" (FGC) dataset, which is a spoken multiple-choice question answering task in Mandarin Chinese, in the experiments. Each passage-question-choices (PQC) set contains a passage, a question, and 4 candidate choices, among which only one choice is the correct answer. The domain of the FGC dataset is highly diverse, including science, news, medicine, literature, history and so on. The training set consists of 7,072 PQC examplers, and there are 1,500 PQC examplers for development. An elementary and an advanced test sets were investigated in this study, and both of them contains 1,000 PQC examplers. It is worthy to mentioned that questions in the advanced test set require deep understandings for choosing correct answers.

### 4.2. Acoustic Features & ASR System

Our ASR system was built up using the Kaldi toolkit [28], where the acoustic model was trained based on TDNN-F with lattice-free MMI [29, 30], followed by model refinement with sMBR [31], with

---

[1] Formosa Grand Challenge - Talk to AI: https://fgc.stpi.narl.org.tw/activity/techai2018

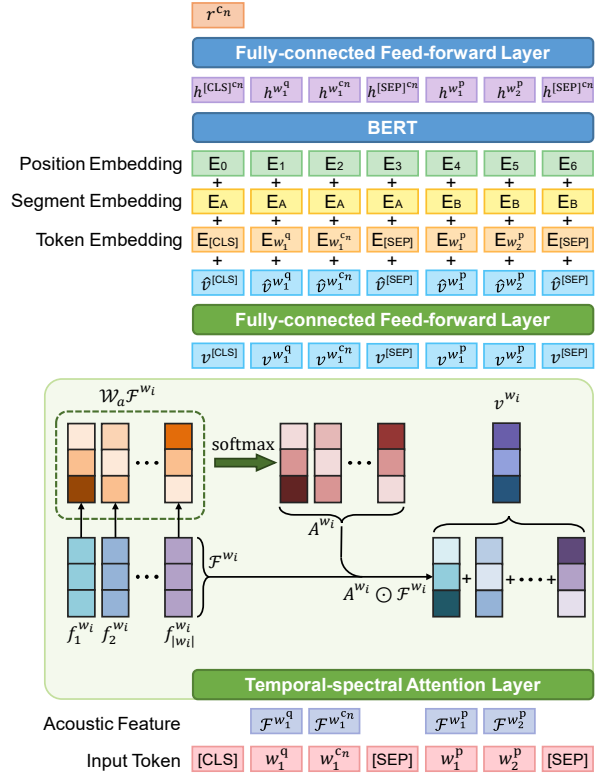

Figure 1: *An overview of the proposed Audio-enriched BERT-based (aeBERT) framework. Note that $v^{[CLS]}$ and $v^{[SEP]}$ are set to zero vector.*

487 hours of TV and radio broadcasting speech. In audio processing, spectral analysis was applied to a 25 ms frame of speech waveform every 10 ms. For each acoustic frame, 40 MFCCs derived from 40 FBANKs, plus 3 pitch features, were used for ASR and for our proposed aeBERT framework. Utterance-based mean subtraction was applied to these features. The lexicon contained 91,573 Chinese words. The word-based trigram language model was trained with Kneser-Ney backoff smoothing using the SRILM toolkit [32]. The recurrent neural network language model was used for lattice rescoring [33]. The training corpus was compiled from PTT[2] articles (2018) and CNA news stories (2006~2010) [34]. The character error rate of our ASR system is about 7.79%.

### 4.3. Implementation Details

The BERT model (bert-base-chinese) used in this paper is implemented by the Huggingface's Transformers library [35]. For better initializing TSAtt, a pretraining process is performed for one epoch, and the objective is to minimize the mean square error between the generated acoustic-level representation (i.e., $v^{w_i}$, c.f. Section 3.2) and the token embedding from BERT for each token in the training examplers. To that end, TSAtt seems to be more effective to learn a complementary embedding from a set of acoustic feature vectors to the conventional text-level embeddings. Moreover, to warm up the finetuning process, we first adapt the model parameters with auto-transcribed text only for an epoch. After that, we completely finetune all the model parameters with both auto-transcribed texts and acoustic feature vectors. The framework was implemented by PyTorch [36], and the model parameters were optimized by the AdamW method [37]. The learning rate is set to 0.001 for TSAtt pretraining, and 3e-5 for the rests. The batch size is set to 2, and the gradient is accumulated 32 times for each update.

---

[2] PTT: https://www.ptt.cc/index.html

## 5. Experimental Results

In the first set of experiments, we evaluate various baseline systems, and the results are listed in Table 1. Systems compared in this study include a naïve baseline, a word embedding-based method and three recently proposed neural-based SOTA MCQA methods. The most naïve baseline is to choose the longest choice or the shortest choice as the answer (denoted by "Choice Length"). This method could be even worse than a random guess. In addition to the naïve baseline system, a simple strategy based on the word embeddings is investigated. The method employs the pretrained word embeddings to represent a passage/question/choice by averaging the embeddings of all the words in the passage/question/choice. Then, we can choose the choice with the largest cosine similarity with the passage or the question to be the answer. The word embeddings used in this study are trained by fasttext [38] on the same corpus for ASR language model training (c.f. Section 4.2). The dimension of the word embedding was set to 300. The results, as denoted by "Choice Similarity" in Table 1, indicate that the relationship between the question and the choice is more effective than the relationship between the passage and the choice. Next, the recently proposed neural-based methods are also compared in this study, including the QACNN [13], the HMM [22], and the vanilla BERT (c.f. Section 3.1). It is worthy to note that both the QACNN and the vanilla BERT models only leverage auto-transcribed text for answer prediction, while the HMM uses both text-level and acoustic-level information for the SMCQA task. Valuable observations can be drawn from the results. First, as expected, the QACNN, the HMM and the vanilla BERT models performed much better than the "Choice Length" and "Choice Similarity" methods, which also reveal the ability and the potential of the neural-based methods for SMCQA task. Next, we can observe that the HMM outperforms QACNN in all cases. The reason should be that the HMM model integrates both text-level and acoustic-level information for answer prediction, while the QACNN model only leverages the text-level information for SMCQA task. Moreover, the vanilla BERT can achieve the best results than all of the other baselines, which witnesses again the giant successes of the research on language representations.

In the second set of experiments, we make a step forward to compare the proposed aeBERT with all of the baseline systems, and the experimental results are also presented in Table 1. Based on the results, several worthwhile observations can be made from the comparisons. At first glance, we find that the proposed aeBERT outperforms all of the baseline systems in all cases, which signals that it can indeed make use of both acoustic-level and text-level statistics in a systematic and theoretical way for SMCQA answer prediction. Second, TSAtt pretraining (c.f. Section 4.3) is definitely an important step, since results for aeBERT are better than results for aeBERT(w/o TSAtt Pretrain). Third, when looked into the table, results for the advanced test set are worse than for the elementary test set in almost all cases, which reveal that questions in the advanced test set require deep understandings for choosing correct answers. Fourth, because of the BERT model, the vanilla BERT and the proposed aeBERT models can absolutely outperform other neural-based methods (i.e., the QACNN and the HMM), especially in the test sets. To sum up, by manipulating both text-level and acoustic-level information, the proposed aeBERT framework is the affirmative choice for the SMCQA task.

Table 1: *Performance (in accuracy (%)) of different systems.*

| Model | Dev | Test | |
|---|---|---|---|
| | | Elementary | Advanced |
| Choice Length | | | |
| Longest | 39.14 | 29.58 | 32.36 |
| Shortest | 19.74 | 23.23 | 20.94 |
| Choice Similarity | | | |
| Passage-Choice | 26.25 | 25.08 | 24.95 |
| Question-Choice | 47.09 | 38.08 | 35.07 |
| QACNN | 63.12 | 71.23 | 39.07 |
| HMM | 66.14 | 72.00 | 40.98 |
| Vanilla BERT | 68.07 | 77.00 | 47.90 |
| aeBERT | | | |
| w/ TSAtt Pretraining | **70.53** | **79.20** | **49.50** |
| w/o TSAtt Pretraining | 69.13 | 79.10 | 49.20 |

Table 2: *Performance (in accuracy (%)) of the vanilla BERT method with respect to manual transcriptions (Manual) or auto-transcribed texts (ASR).*

| Data Usage | | Experimental Result | | |
|---|---|---|---|---|
| Training | Dev & Test | Dev | Test | |
| | | | Elementary | Advanced |
| Manual | Manual | 80.80 | 91.00 | 56.30 |
| ASR | Manual | 67.73 | 77.80 | 48.80 |
| ASR | ASR | 68.07 | 77.00 | 47.90 |

At the last stage, in order to exam the effect caused by the ASR errors for SMCQA task, we take vanilla BERT, which only considers text-level information for answer prediction, as a subject. As the upper bound, the vanilla BERT model is trained on manual-transcribed PQC sets, and the development and test sets are also in the form of manual-transcribed text. Orthogonal to the upper bound system, a model trained with erroneous transcripts by ASR is obtained, and the performances of the MCQA task in either with and without recognition errors are evaluated. All of the results are summarized in Table 2. The results indicate a significant performance gap between the upper bound system and other settings, which shows that the recognition errors inevitably mislead the predictions for the vanilla BERT model so as to degrade the MCQA performance. Accordingly, the analysis implicitly suggests that extra information, besides auto-transcribed text, should be explored to improve the SMCQA system. In summary, the proposed aeBERT is deemed a preferable vehicle for utilizing acoustic-level and text-level characteristics in the SMCQA task.

## 6. Conclusion

In this paper, we have presented an audio-enriched BERT-based (aeBERT) framework, which jointly considers the acoustic-level and text-level statistics for the SMCQA task. The proposed framework has been evaluated on the 2018 Formosa Grand Challenge (FGC) dataset. The experimental results demonstrate its remarkable superiority than other strong baselines compared in the paper, thereby indicating the potential of the framework. For future work, we plan to extend the proposed aeBERT to other NLP-related tasks, such as retrieval and summarization, and evaluate the framework on other spoken question answering datasets.